\documentclass[letterpaper, 10 pt, conference]{Resources/IEEEconf}  

\IEEEoverridecommandlockouts                              

\usepackage{etoolbox}
\makeatletter
\patchcmd{\@makecaption}
  {\scshape}
  {}
  {}
  {}
\makeatletter
\patchcmd{\@makecaption}
  {\\}
  {.\ }
  {}
  {}
\makeatother

\usepackage{graphicx}
\let\labelindent\relax 
\usepackage{lipsum}  
\usepackage{hyperref}
\usepackage{enumitem}
\usepackage{xcolor}
\usepackage{array}
\newcolumntype{x}[1]{>{\centering\let\newline\\\arraybackslash\hspace{0pt}}p{#1}} 
\usepackage{svg}
\usepackage{amsmath}
\svgpath{{../Figures/}}
\usepackage{float} 
\usepackage{algorithm}
\usepackage{wrapfig}
\usepackage[noend]{algpseudocode}

\title{\LARGE \bf
The Door and Drawer Reset Mechanisms: Automated Mechanisms for Testing and Data Collection
}
\author{Kyle DuFrene\textsuperscript{1}, Luke Strohbehn\textsuperscript{1}, Keegan Nave\textsuperscript{1}, Ravi Balasubramanian\textsuperscript{1}, and Cindy Grimm\textsuperscript{1}
\thanks{\textsuperscript{1}Collaborative Robotics and Intelligent Systems (CoRIS) Institute, Oregon State University, Corvallis, OR 97331 
\newline\indent Thank you to our partners at the UMass Lowell Nerve Center, Brian Flynn, Adam Norton, and Holly Yanco.
\newline\indent This work was supported in part by the NSF under Grants CCRI 1925715 and RI 1911050.}}

\begin{document}
\maketitle
\thispagestyle{empty}
\pagestyle{empty}

\begin{abstract}
Robotic manipulation in human environments is a challenging problem for researchers and industry alike. In particular, opening doors/drawers can be challenging for robots, as the size, shape, actuation and required force is variable. Because of this, it can be difficult to collect large real-world datasets and to benchmark different control algorithms on the same hardware.  In this paper we present two automated testbeds, the Door Reset Mechanism (DORM) and Drawer Reset Mechanism (DWRM), for the purpose of real world testing and data collection. These devices are low-cost, are sensorized, operate with customized variable resistance, and come with open source software. Additionally, we provide a dataset of over 600 grasps using the DORM and DWRM. We use this dataset to highlight how much variability can exist even with the ``same'' trial on the ``same'' hardware. This data can also serve as a source for real-world noise in simulation environments.
\end{abstract}

\section{Introduction}\label{section:intro}

Incorporating robots in human environments is a challenging problem that requires the fusion of many different disciplines. Robots for tasks such as mopping and vacuuming have successfully been integrated into many homes, but tasks that require general manipulation have yet to see the same level of success. Actions that require opening doors and drawers can be particularly challenging for robots due to door/drawer variation and the required manipulation fidelity.

In the past few decades, significant progress has been made in control models, such as impedance control, for opening doors and drawers~\cite{adaptive_control}. However, this approach requires accurate force-torque sensors located in the manipulator or end effector. Recently, learning methods are becoming more commonplace in grasping and manipulation, and have been applied to door opening~\cite{mochurad_approach_2023, urakami_doorgym_2022}. These machine learning methods require large amounts of data to train on. Given the varying geometry and force required to open different doors and drawers, training must also include a diverse set of mechanisms~\cite{review}. Because of this, many machine learning approaches use simulated data to train. This, unfortunately, leads to a loss of real world noise and fidelity in their datasets~\cite{simtoreal, review}. 
\begin{figure}
    \centering
    \includegraphics[width=0.48\textwidth]{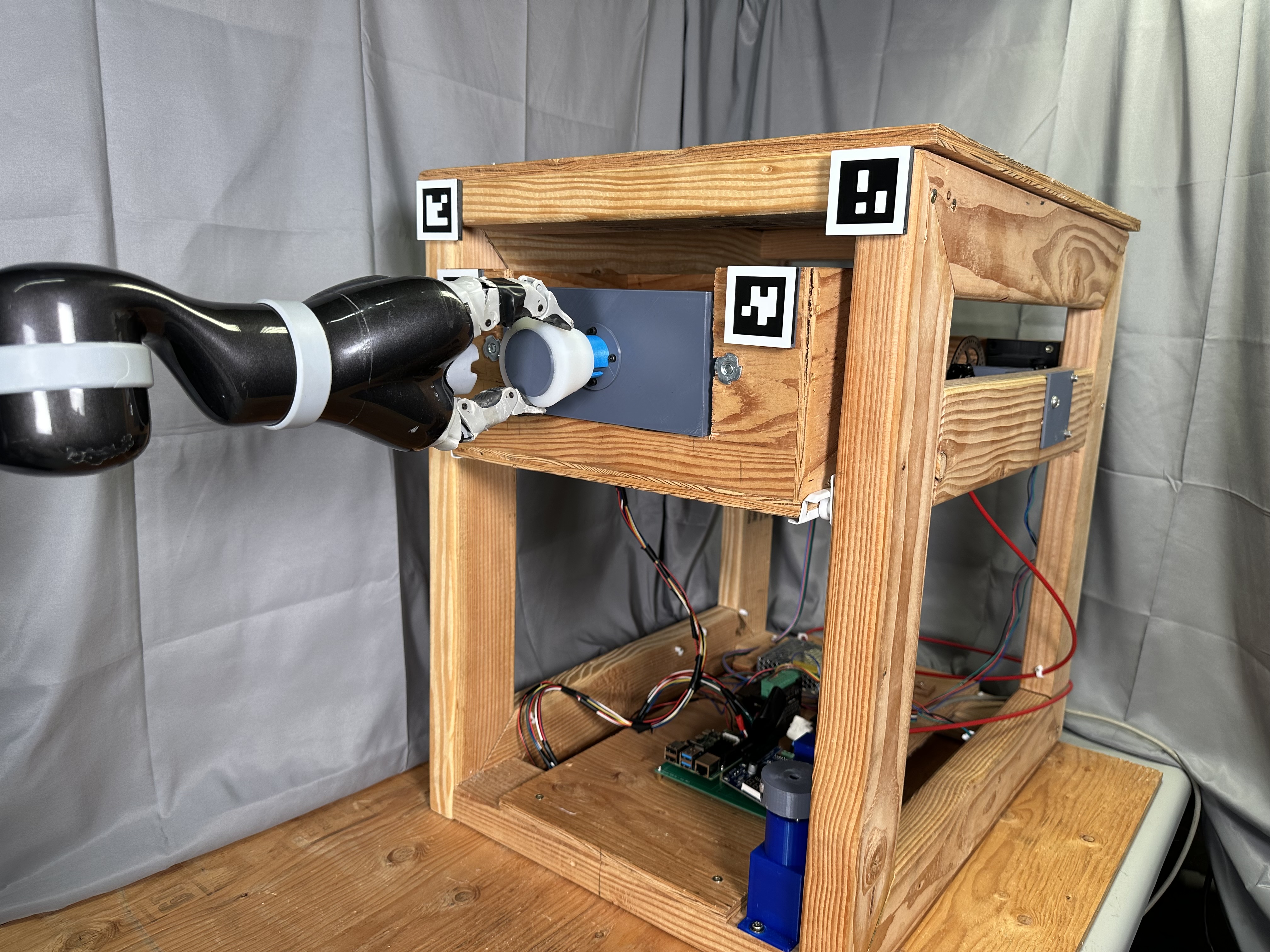}
    \caption{The Drawer Reset Mechanism with a Kinova JACO 7-dof arm. As pictured, the instrumented knob is attached to the Drawer Reset Mechanism.}
    \label{fig:overall_arm_drawer}
\end{figure}

To improve robotic manipulation of doors and drawers, there is a need for repeatable testing mechanisms for benchmarking control algorithms, and generating datasets. To the best of the authors' knowledge, there are no current open-source door and drawer testbeds and accompanying software packages for this end. In this paper, we propose two testbeds for the collection of data and benchmarking of door and drawer manipulation tasks, as well as the accompanying software and datasets.

The Door Reset Mechanism (DORM) is an artificial door which automatically closes, interfaces with a variety of pull attachments (handles/knobs), and provides variable resistive forces against opening. Additionally, the DORM has a variety of sensors to provide force feedback and door angle. The Drawer Reset Mechanism (DWRM) provides all the functionality of the DORM, except in the form of a drawer. Each testbed is low cost (under \$400), highly customizable, and sensorized. The built in sensors, combined with external cameras and data from the manipulator, allow us to quickly create robust datasets or compare proposed control methods. Each testbed can record, reset, and be monitored remotely, making large dataset collection feasible and repeatable.  

Paired with the testbeds, our modular software packages perform automatic data collection and processing, and are compatible with most robotic manipulators and standard visualization tools, such as RViz~\cite{rviz}. 
\begin{figure*}
    \centering    \includegraphics[width=.98\textwidth]{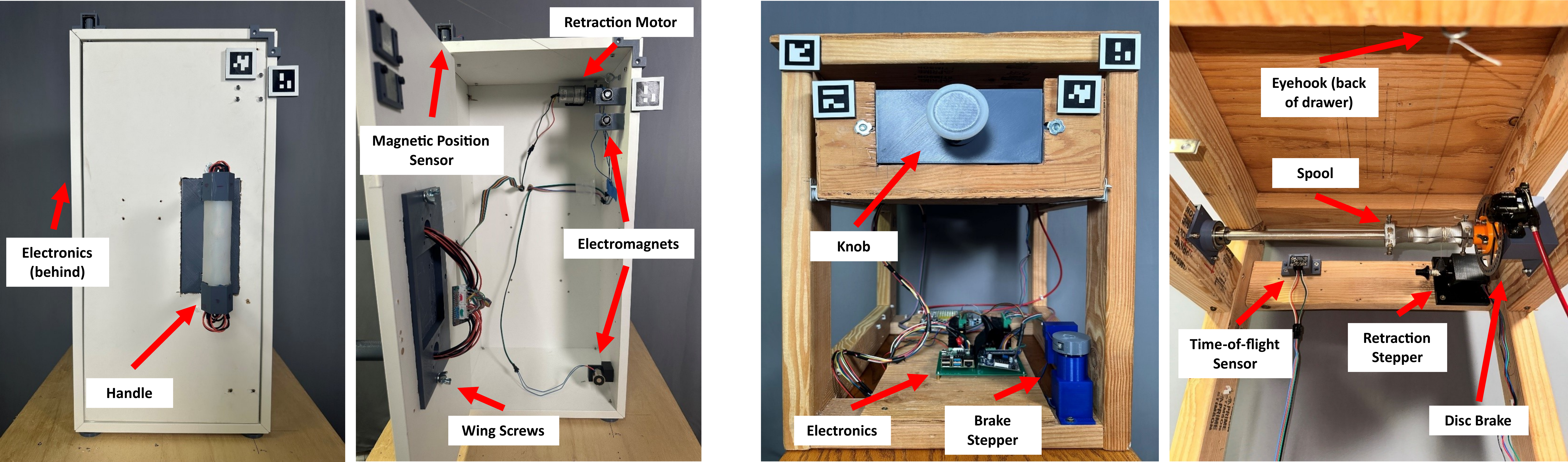}
    \caption{Pictured on the left is the Door Reset Mechanism, on the right is the Drawer Reset Mechanism. Important parts are labeled, including electronics, sensors, motors, and pull attachments.}
    \label{fig:door_drawer_diagrams}
\end{figure*}

This work presents the design of the testbeds, associated software, and a dataset to demonstrate the capabilities of the testbeds. With the DORM and DWRM, we conducted a total of 660 trials with a Kinova Gen 3 robot arm and a Kinova JACO robot arm. Trials were conducted with an instrumented handle and knob, a variety of resistance forces (applied by the DORM and DWRM), and multiple grasp types. The dataset includes video, manipulator joint states, instrumented pull attachment forces, and trial result for each test.

\noindent In this work our contributions include: 
\begin{itemize}
   \item Instrumented hardware platforms for automated trials of opening doors and drawers, the Door Reset Mechanism (DORM) and Drawer Reset Mechanism (DWRM).
   \item Software packages for conducing grasp trials and visualization.
   \item A dataset of over 600 grasps.
\end{itemize}

\begin{table*} 
\caption{Design considerations for the Door and Drawer Reset Mechanisms, and their corresponding solutions.}
\begin{center}
\renewcommand{\arraystretch}{1.5}
\begin{tabular}{ |p{5.5cm}|p{11cm}|  }
 \hline
 Design Requirements& Design Solutions\\
 \hline
 Cost   & The DWRM frame is built from 2x4s and plywood, and the DORM from an Ikea cabinet. Custom parts are either 3D printed (PLA), or available at common online retailers. Each can be build for under \$400.\\
 
 Replicability   & All components are readily available at retailers or 3D printed. 3D CAD, wiring diagrams, and software are all provided on the website. \\
 
 Compatibility with a wide range of manipulators& Manipulators are prompted by a ROS action client. Any manipulator that be controlled within a custom ROS action server is compatible.\\
 
 Robust and accurate resetting (closing) & Simple pulley systems with high-tensile strength string. Error checking with position/orientation sensors. \\
 Variable resistance to opening& A disk brake on the DWRM provides variable resistance between $0$ and $25$ Newtons of resistance. Electromagenets in the DORM provide between $0$ and $10$ Newtons of resistance.\\
 
 Position/orientation feedback& A time-of-flight sensor inside the DWRM measures drawer position and a magnetic rotatory position sensor above the DORM measures orientation of the door.\\
 
 Compatibility with various pull attachments&Any pull attachment that can be 3D printed, or attached to a 3D printed base is compatible. The base is attached to the DORM/DWRM with two thumb screws.\\
 
 Force feedback on pull attachments& The handle has 12 evenly distributed force-sensitive resistors, the knob has 5. While not precise, they provide accurate models of contact forces on the pull attachments.\\
 
 \hline
\end{tabular}
\end{center}
\label{table:design}
\end{table*}

All aspects of the testbeds' designs and dataset discussed in this paper are open source and are available at \href{https://osurobotics.github.io/Physical-Robotic-Manipulation-Test-Facility/}{https://osurobotics.github.io/Physical-Robotic-Manipulation-Test-Facility/}.

\section{Related Work}
Due to the challenges of real world data collection and testing, many existing benchmarks and testing environments are provided through simulation such as~\cite{rlbench, urakami_doorgym_2022, zhu_robosuite_2022}. Work from Urakami, et al. demonstrates an open source randomized environment for door opening tasks as well as a baseline agent for benchmarking purposes~\cite{urakami_doorgym_2022}. However when transferred to the real world, the authors' note performance degradation. Similarly, in~\cite{ding_sim--real_2021} tactile sensing is used during training of their model, and although their sim-to-real performance is improved, there still exists a gap in performance. 

When transferring from simulation to the real world, door and drawer opening tasks require some form of mechanism for benchmarking. Karayiannidis et al. and  Mochurad et al. used random real doors to benchmark their control approaches~\cite{adaptive_control,mochurad_approach_2023}. In both papers, a manual measurement of door angle was required to determine if a trial was successful or unsuccessful. In other work such as~\cite{gu_deep_2017} small mock doors with built-in IMU sensors were used to test and train their policy in real world trials. However due to constraints on their real world testbed they were unable to complete all of the tasks done in simulation. Across all work, there is no definitive benchmarking tool, and the definition of success can vary from paper to paper.

General grasping and manipulation datasets such as~\cite{google} and~\cite{supersizing} are useful for training models, however to the best of the authors' knowledge, there are no existing open source, real world grasp datasets for door and drawer grasping/manipulation.

\section{The Mechanisms}

The Door and Drawer Rest Mechanisms are designed to automate the resetting (closing) of a door and a drawer for grasp and pull testing. As part of the resetting process, the testbeds return to their closed state and resistance forces are set. To support many manipulators, we use a state-machine based interface that triggers resetting, manipulator control, and data-collection actions. The testbeds also have sensing built-in to collect information on how fast/far they were opened and where on the handle the grasp applied force. Design requirements for these features and corresponding solutions are presented in Table~\ref{table:design}. 

The entire system is broken into into three components: the mechanical design of the two testbeds, electrical design, and software.
\subsection{Mechanical Design}

The Door and Drawer Reset Mechanisms provide the same functionality in slightly different ways. Both require a certain amount of force to be opened, and both return to the original state after the grasp and open operations have completed. Additionally, each testbed interfaces with a modular pull attachment such as a handle or knob. 

\subsubsection{Door}
The Door Reset Mechanism is built from a modified EKET cabinet from IKEA. As the door may experience more extreme forces that it was designed for, 3D printed braces placed in the corners maintain structural integrity. The door opens and closes as usual about the hinge location. A high tensile strength braided Kevlar string is attached to the opposite side of the door and connected to a 12 volt DC gearmotor attached to the frame. The string is wound around a spool, which is used to retract the string and close the door. Before a trial, the sting is spooled out a few inches by the motor with a small fishing weight maintaining minimal tension. 

Three electromagnets mounted to the frame provide variable resistance to the door opening. Combined, they can provide a resistance of up to 10 Newtons. 

The top hinge replaced by a 3D printed hinge with a small magnet mounted in the top. Above the magnet, an AS5047P magnetic rotatory position sensor is mounted to the frame. The sensor detects the rotation angle of the magnet, providing precise orientation data of the door.

A cut-out in the door allows for mounting of various pull attachments, as discussed later. Additionally, a custom PCB is mounted behind the DORM, and cut-outs in the rear allow wires to pass inside.

To protect the DORM in event of extreme manipulator movements, the base of the DORM mounts to the table with magnetic feet. The feet resist larger amounts of forces in the forward and sideways directions, but forces backwards (into the door) and excessive forces will dislodge the feet from the table. This prevents the DORM from tipping over or otherwise being damaged during extreme movements.

Figure~\ref{fig:door_drawer_diagrams} presents a labeled diagram of the DORM.

\subsubsection{Drawer}
The Drawer Reset Mechanism's frame is custom assembly of wood 2x4s and plywood. Movement of the drawer occurs along drawer slides mounted to the bottom edge of the box. An L bracket is fixed to the frame of the reset structure to ensure a full stop of the drawer in case of mechanical failure. An eye ring screw is fixed to the back of the drawer to allow the reset mechanism to pull the drawer back to the closed state via a high tensile-strength braided Kevlar string.

A 1/2" aluminum axle is located behind the drawer, and serves as the main component of the reset mechanism. The axle is attached to bearings on both sides, which are then physically pressed into custom 3D printed mounts bolted to the frame. A 3D printed double spool is fixed to this axle, where one side of the spool winds the string attached to the drawer. This string remains taut at all times to prevent tangles and knots. The other side of the spool winds an identical string in the opposite direction, which then winds around a spool attached to a NEMA 17 stepper motor. Similar to the door, this string is unwound before trials to allow the drawer to open, and minimal tension is maintained with a small fishing weight.

Next to the double spool and also fixed to the aluminum axle is a disc brake, fixed with custom 3D printed mounts. The brake is screwed to one of the custom 3D printed bearing mounts. This brake allows for variable resistance to be applied during a drawer pull operation, and is subsequently released when the reset operation occurs.

A VL53L0X time-of-flight (ToF) sensor is fixed to the back of the drawer structure and points directly at the back of the drawer for high-accuracy measurements of drawer location.

The braking force is applied to the disc by pulling the braking cable with a 100:1 geared NEMA 17 stepper motor. A 3D printed brake cable housing is mounted to the stepper to secure the cable.

During a reset operation, the reset string is wound in towards the stepper and away from the axle spool. In a similar but opposite manner, the spool connected to the drawer simultaneously winds in, pulling the drawer back to its closed state. The reset stepper continues to wind in until the ToF sensor returns a previously-determined minimum distance value of the drawer. Once stopped, the reset stepper then reverses direction and unwinds the string, with a fishing line weight providing minimal tension to prevent tangles. Introducing slack into the reset string allows subsequent drawer pulls by the robotic arms so that they only feel the resistance of the applied brake and the weight of the drawer itself, without the interference of the reset stepper motor.

Figure~\ref{fig:door_drawer_diagrams} presents a labeled diagram of the DWRM.

\subsubsection{Pull Attachments}
\begin{figure}
    \centering
    \includegraphics[width=0.48\textwidth]{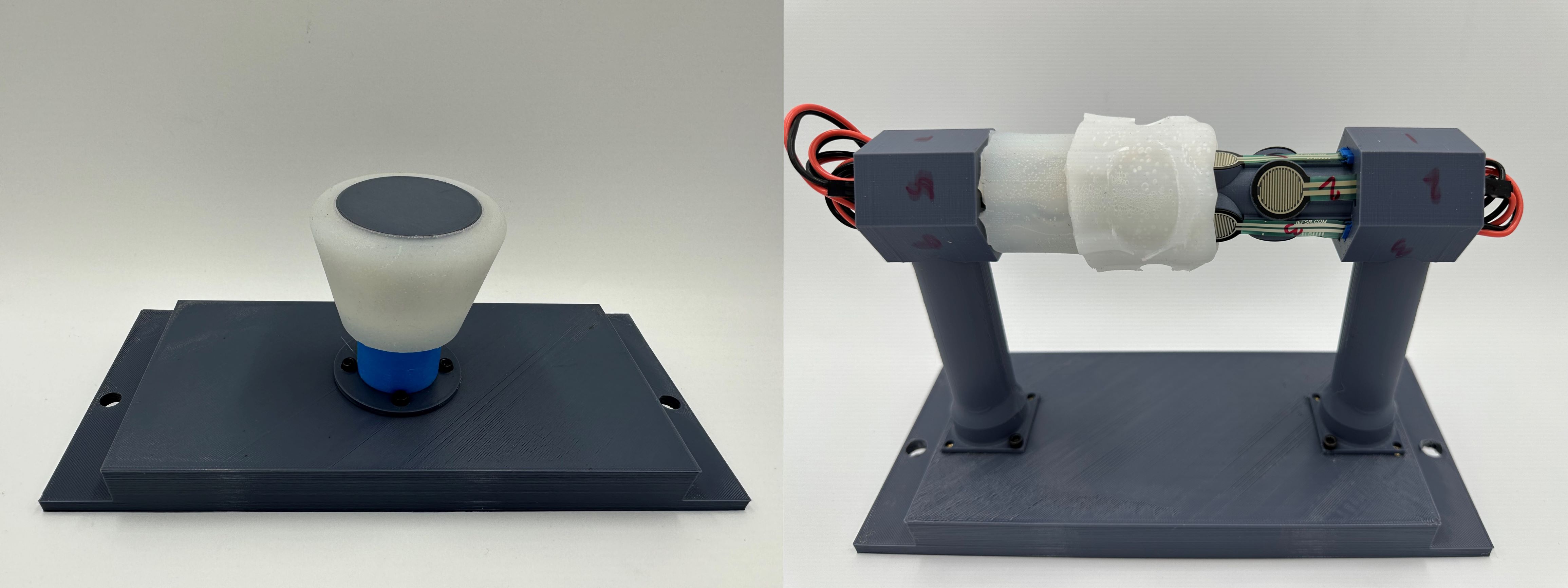}
    \caption{The knob (left) and handle (right) used as the pull attachments for the dataset in this paper. Each have a silicone wrapper to protect and distribute grasping forces to the force sensitive resistors (FSRs). The silicone on the handle is rolled back to show the FSRs.}
    \label{fig:handle_knob}
\end{figure}
Both the door and drawer accept the same pull attachments. For this paper we focus on a handle and a knob (shown in Figure~\ref{fig:handle_knob}), however the attachment may be customized to be any shape and vary in size. Two thumbscrews mount the pull attachment to the testbed, allowing for easy installation and removal.

The handle mimics a handle that may be found on an entry door to a commercial building. The handle is 3D printed, and has a layer of twelve separate 0.5" force sensitive resistors (FSR) evenly distributed across the surface. A layer of silicone is wrapped around the handle to protect the FSRs and distribute forces. 

The knob is similar in size and geometry to a doorknob in a standard residential home. Similar to the handle, it is 3D printed and has a layer of five FSRs placed around the knob. A layer of silicone protects the FSRs and distributes forces.
\subsection{Electrical Design}
The Door and Drawer Reset Mechanisms feature custom-designed printable circuit boards (PCBs) for an organized layout and removable electrical connections. Power is provided to the PCBs via 5 and 12 volt power supplies. A 3.3 volt buck converter steps down the 12 volt input power to use with the pull attachment sensors. All connections to and from the PCBs are either JST type plugs or screw-terminals for quick and reliable connections. 

Both PCBs use the same setup for pull attachment FSRs. Two analog to digital converters (ADCs) are located on the PCB and connected to the Raspberry Pi with serial communication. The ADCs analog pins are connected to a 12 pin JST plug, which ultimately connects to the FSRs through pull-down resistors. A 14 pin cable runs from the 12 pin JST plug and two screw terminals (for power and ground) to a small breakout PCB located on the pull attachment itself. There, 10 kiloOhm pull-down resistors pull the data pins to ground. A data pin and voltage is connected to each FSR. 

Specific to the DORM, the rotatory magnetic position sensor is connected to the Raspberry Pi via serial peripheral interface (SPI). Additionally, an H-bridge motor driver and transistor drive the DC motor and electromagnets, respectively.

For the DWRM, the ToF communicates with the Raspberry Pi via I\textsuperscript{2}C. Both stepper motors are driven by microstepper drivers. The microstepper drivers are connected to general purpose input output (GPIO) pins on the Raspberry Pi.

\subsection{Software}

\begin{figure}
    \centering
    \includegraphics[width=0.48\textwidth]{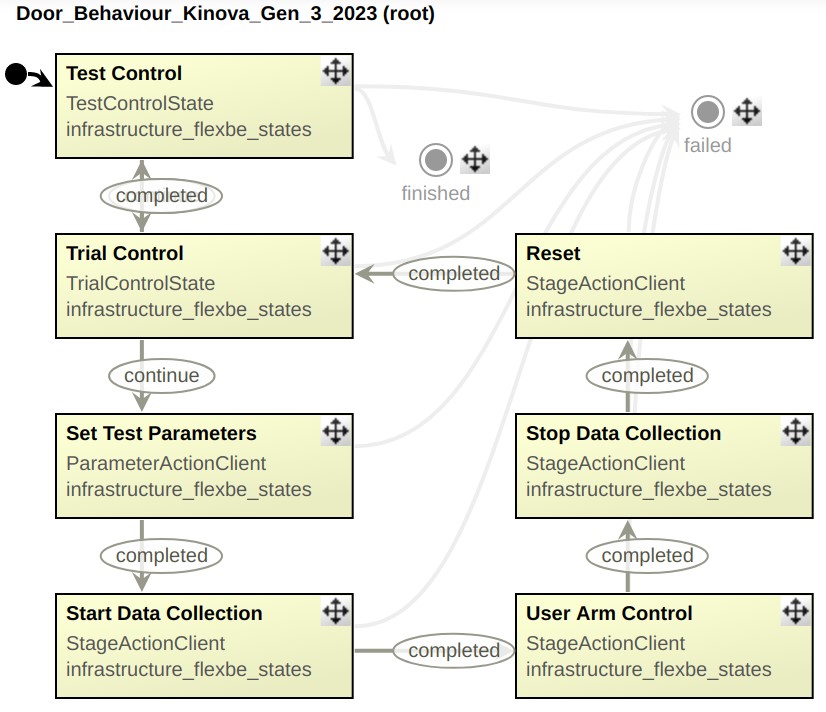}
    \caption{The FlexBE state machine showing trial progression for standard grasping trials with the DORM and DWRM.}
    \label{fig:flexbe}
\end{figure}

The software used for both Door and Drawer Reset Mechanisms builds off of previous work in~\cite{grasp_reset}, and uses the Robot Operating system (ROS) for communication and data collection. Each testbed has an onboard Raspberry Pi which communicates with a central control computer using ROS nodes. Video and sensor data is automatically recorded and labelled with trial metadata after each run. To trigger data collection, arm code, and closing operations, the state machine FlexBE~\cite{flexbe} is used. This software framework allows us to easily switch between arms, grippers, and testbeds.

\begin{figure*}
    \centering    
    \includegraphics[width=.8\textwidth]{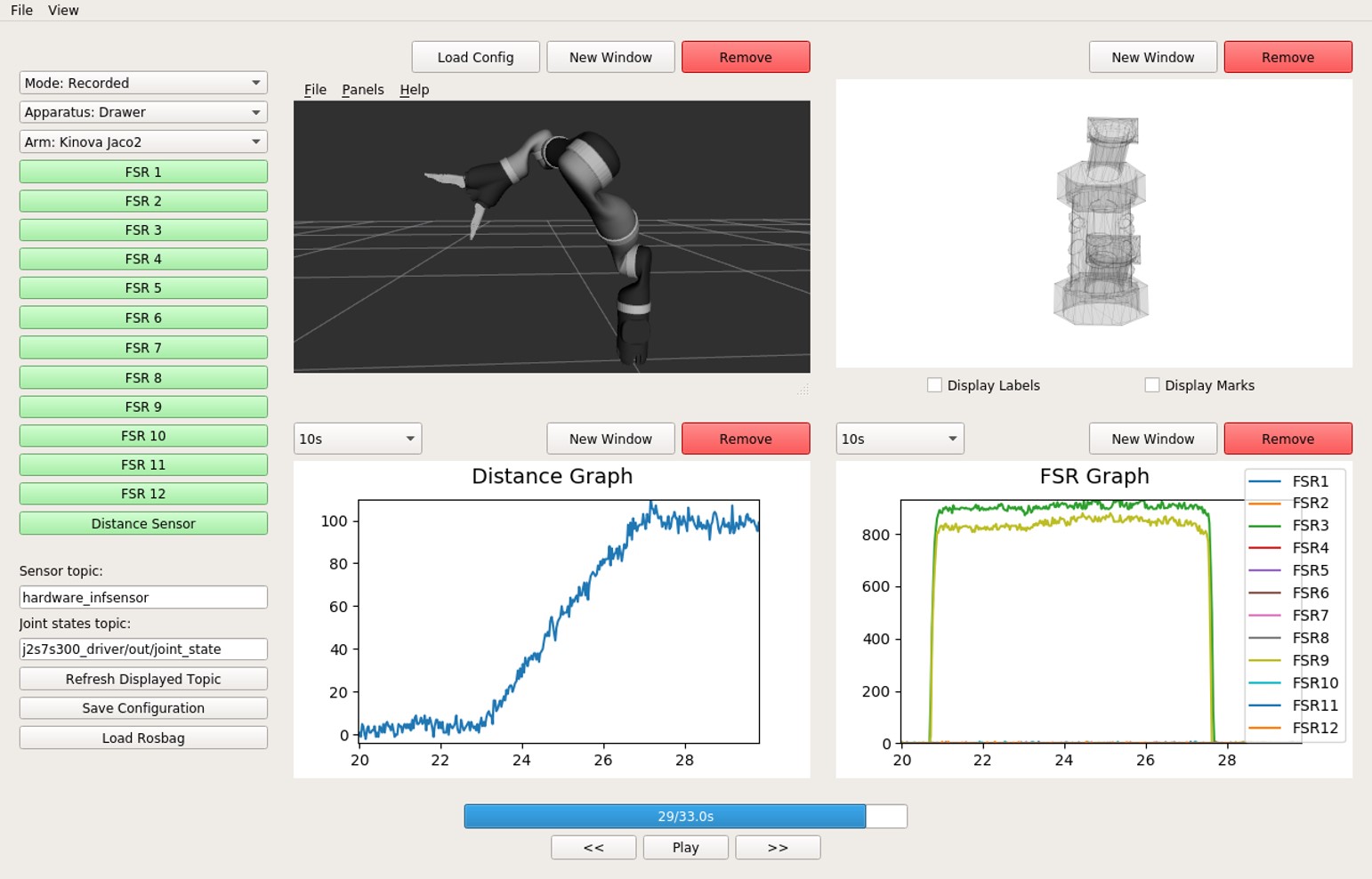}
    \caption{Our visualisation interface showing a trial. Buttons, drop-downs, and text boxes allow user inputs to enable or disable certain features. Four visualization windows are displayed. Top left: a view of the Kinova JACO 2 states, Top right: a 3D model of the handle, Bottom left: a graph of the drawer distance recorded by the ToF, Bottom right: a plot of FSR data.}
    \label{fig:visual}
\end{figure*}

In addition to our core control software, we also include a visualization tool. This allows users to watch sensor data live, or play back recorded rosbag data from any trial. Visualization includes: arm path reconstruction using an embedded RViz window, sensor plots, and 3D models for activated FSR's on the pull attachment. The visualization tool is modular --- it can be expanded to support addition testbeds, manipulators, and sensors. A view of the interface with an trial from the dataset is shown in Figure~\ref{fig:visual}.

All software and related documentation is available at our website provided in Section~\ref{section:intro}, and an in-depth look at the software design can be found in our related work~\cite{grasp_reset}. 

\section{Dataset}\label{section:dataset}
In addition to the physical testbeds, we present a dataset with multiple manipulators, pull attachments, grasps, and resistances.

\subsection{Dataset Methods}
Our goal was to create a dataset of grasps and pulls under diverse conditions to represent real-world variation and noise. This dataset demonstrates the capabilities of the testbeds, and can be used for benchmarking or learning purposes.

The dataset consists of multiple grasps, with both pull attachments, and a variety of resistances on both the DORM and DWRM. For the DORM, all trials were conducted with a Kinova Gen3 paired with a RobotIQ 2F-85 gripper. For the DWRM, all trials were conducted with a Kinova JACO arm with its standard 3-finger gripper. A high-level view of the combinations is presented in Table~\ref{tab:dataset}.

\begin{table*}[]
    \caption{All combinations of grasps presented in the dataset.}
    \centering
    \def\arraystretch{1.4}%
    \begin{tabular}{|m{5.5em}|m{13em}|m{6.75em}|m{6em}|m{9em}|m{4em}|}
    
    \hline
      \textbf{Mechanism} & \textbf{Manipulator} & \textbf{Handle Grasps} &  \textbf{Knob Grasps} & \textbf{Resistances} & \textbf{\# Trials}\\
      \hline
      Door & Kinova Gen3 + RobotIQ 2F-85 & 5 & 5 & 0, 5, 10 N & 10\\
      \hline
      Drawer & Kinova JACO 2 & 4 & 2 & 0, 7, 10, 15, 20, 25 N & 10 \\
      \hline
    \end{tabular}

    \label{tab:dataset}
\end{table*}
\begin{figure*}
    \centering
    \includegraphics[width=.98\textwidth]{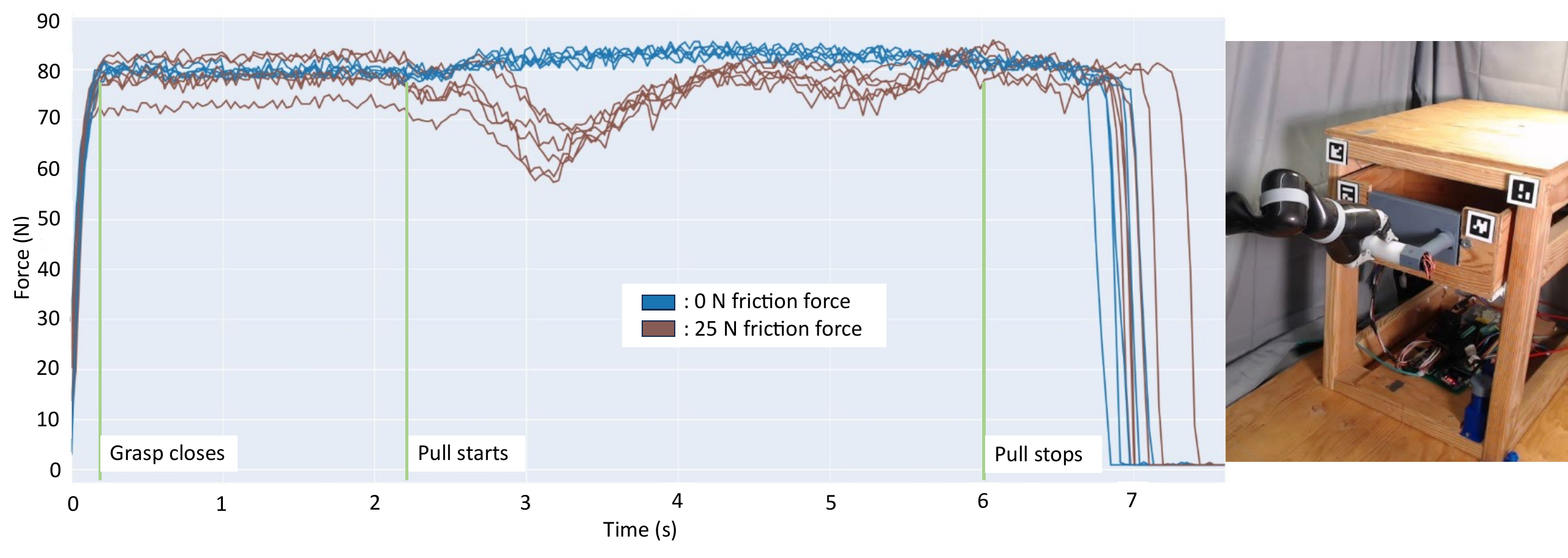}
    \caption{A plot showing the force recorded by FSR 9 on the handle for multiple 0 and 25 Newton trials of the Drawer Reset Mechanism over the duration of the pull operation. Of the 10 repeating trials, trials with significantly longer pull times or outlier force values were filtered out. The rest of the trials were normalized over time to the start of a sharp rise (grasp) and a sharp fall (grasp release). FSR 9 is directly under the top right finger of the Kinova JACO 2 gripper. A picture of one of the trials in progress is shown to the right.} 
    \label{fig:fsr9}
\end{figure*}

We selected grasps that represent various common ways a door/drawer may be opened. For example, for the DORM with the knob, we preformed the following five grasps:
\begin{itemize}
    \item Palm horizontal grasp (straight on)
    \item Fingertip horizontal grasp (straight on)
    \item Top down horizontal grasp
    \item Fingertip angled grasp
    \item Fingertip vertical grasp
\end{itemize}

Due to grasping force exerted by the RobotIQ gripper for the DORM trials, we only test successful grasps. For the DWRM, the Kinova JACO fingers are underactuated and have significantly less grasping power. We focused on sampling successful grasps, however there were some failures.

\subsection{Dataset Results}
A total of 660 trials were conducted, 300 with the DORM and 360 with the DWRM. All trials on the DORM were successful, but a small number of fingertip grasps combined with higher resistance forces applied by the DWRM were failures. 

As we collected the dataset in the real world, our dataset includes real world noise and inconsistencies that are difficult to represent in simulation. For example, each trial is individually planned with MoveIt, and varies based on starting position due to tolerances in the joints. As a result, each manipulator path may have slight variation, and the actual manipulator (especially the Kinova JACO 2) follows the path with some deviation. Additionally, the force applied by the disk brake in the DWRM varies slightly over the course of a trial (similar to how a real world drawer is not perfectly smooth). The grasp, especially at higher resistance forces and with the underactuated Kinova JACO 2 manipulator, may shift slightly over the trial duration. 

An example of the FSR data collected as described in Section~\ref{section:dataset} is visualized in Figure~\ref{fig:fsr9}. This figure explores the difference in grasp quality of the handle on the drawer between two different friction resistances using an identical grasping pose. Notably, most of the force on FSRs is provided simply by grip strength. Higher resistance forces (the brown lines on the plot) actually show lower FSR readings over the course of the pull movement. This is due to the underactuated Kinova JACO 2 gripper slipping on the handle's silicone layer, and the redistribution of forces across the handle due to the silicone.

The dataset includes all valuable data for each trial. This includes the following information:
\begin{itemize}
    \item Grasp pose
    \item Testbed name and pull attachment
    \item Testbed sensor data (position/orientation of door/drawer)
    \item Pull attachment force readings
    \item Manipulator states (position and velocity)
    \item Rear left and right RGB camera feed
    \item Wrist mounted RGBD camera feed (Kinova Gen3 only)
\end{itemize}


\section{Discussion and Conclusions}

In this paper we propose two low-cost, open source testbeds for automated testing and data collection for door and drawer tasks --- the Door Reset Mechanism and Drawer Reset Mechanism. With our real world automated testing mechanisms, we can emulate a variety of door/drawer resistance forces and handle morphologies. The testbeds also capture valuable data, including distance/orientation and forces on the pull attachment. The testbeds are easy to replicate, and cost less than \$400 each. Because of this, the system is a good benchmarking tool for both hardware and control algorithms. In future work we look to create standardized benchmarking tests that can be used alongside our hardware.

Recording data from the testbeds (in addition to manipulator data, video, and more) is possible with our modular software packages and state machine. We also include a visualization interface for real time or recorded visual information about trials. The testbeds and software enable robust, large scale data collection for training learning models or testing algorithms. 

We additionally present a dataset of 660 grasps to demonstrate our testbeds and provide a resource for training and validation to the robotics community. The dataset demonstrates the importance of real-world trials --- friction, uneven hardware movements, grasping contacts, and other real world variations all are difficult to model in their entirety.

All CAD, schematics, and software for the DORM and DWRM, as well as the accompanying dataset can be found on our website (linked in Section~\ref{section:intro}).

\addtolength{\textheight}{-19.75cm}   
\bibliographystyle{IEEEtran}

\end{document}